\documentclass[runningheads]{llncs}
\usepackage[T1]{fontenc}
\usepackage{graphicx,verbatim}
\usepackage{amsmath}
\usepackage{booktabs}
\usepackage{multirow}
\usepackage{multicol}
\usepackage{cleveref}
\usepackage{amssymb}
\usepackage{bbm}

\usepackage{graphicx}
\usepackage{subcaption}
\newcommand{\interpolatecolor}[3]{%
  \ifdim#1pt<#2pt%
    \cellcolor{red!70!yellow!50}%
  \else%
    \ifdim#1pt>#3pt%
      \cellcolor{green!50}%
    \else%
      \pgfmathsetmacro{\percentage}{100*(#1-#2)/(#3-#2)}%
      \cellcolor{red!70!yellow!50!\percentage!green!50}%
    \fi%
  \fi%
}
\newcommand{\card}[1]{\left| #1 \right|}

\usepackage[table]{xcolor}
\usepackage{colortbl}
\usepackage{upgreek}
\begin{document}
\title{Good Enough? An Investigation on the Impact of Label Quality in Large-Scale Medical Datasets}
\titlerunning{An Investigation on the Impact of Label Quality in Large Datasets}

\author{Alexander Jaus\inst{1,2*} \and
Zdravko Marinov\inst{1,2} \and
Simon Reiß\inst{1}
\and%
Constantin Seibold\inst{3} 
\and
Jiale Wei\inst{1,2}
\and Jens Kleesiek\inst{4\dagger}
\and Rainer Stiefelhagen\inst{1\dagger} 
}
%
%
\institute{Karlsruhe Institute of Technology, Germany \\\email{\{firstname.lastname\}@kit.edu} \and
HIDSS4Health - Helmholtz Information and Data Science School for Health, Karlsruhe/Heidelberg, Germany \and 
Diagnostic and Interventional Radiology, Heidelberg University Hospital \\\email{\{firstname.lastname\}@med.uni-heidelberg.de} \and
Institute for AI in Medicine, University Hospital Essen, Essen, Germany
\email{\{firstname.lastname\}@uk-essen.de}
}

\authorrunning{Alexander Jaus et al.}

\renewcommand{\thefootnote}{\fnsymbol{footnote}}

\maketitle              
\begin{abstract}
Manually refining radiological segmentation masks is highly resource-intensive. To determine when this expert commitment is truly justified for the training of segmentation models, we investigate the relationship between label quality and model performance. Expanding beyond models trained directly for inference, we conduct the first study isolating the impact of label quality in pre-training datasets. While high-quality labels remain essential for models proceeding directly to deployment, we find no evidence that strict label quality is crucial for pre-training efficacy. These results question the necessity of exhaustive human-in-the-loop refinement for massive corpora intended for pretraining and suggest that expert effort is more effectively invested in well-curated downstream target datasets.


\end{abstract}
\begingroup
\renewcommand{\thefootnote}{}   
\footnotemark                   
\footnotetext{*Corresponding author (E-mail: alexander.jaus@kit.edu) \newline
$\dagger$ indicates equal supervision } 
\endgroup

\section{Introduction}
Annotating a dataset the size of the recently proposed DAP-Atlas~\cite{jaus2024towards} would require a radiologist to spend over 12,000 hours, assuming an optimistic $10$ minutes per mask. Other large-scale datasets~\cite{wasserthal2023totalsegmentator,qu2024abdomenatlas,li2024how} are no exception. These massive time requirements make manual annotation clearly infeasible at scale, pushing researchers toward alternative strategies to ensure high-quality masks: DAP-Atlas~\cite{jaus2024towards} relies on algorithmic checks and includes radiologists only for validation, TotalSegmentator~\cite{wasserthal2023totalsegmentator} employs a human-in-the-loop refinement, and AbdomenAtlas~\cite{qu2024abdomenatlas,li2024how} uses uncertainty-based guidance to identify automatically generated masks in need of corrections. This marks a shift in how large datasets are created, where the common approach has shifted from manually annotating masks to refining automatically generated masks. While this significantly reduces manual workload, it still requires substantial radiologist involvement, with dataset curation still taking weeks of radiologist involvement~\cite{qu2024abdomenatlas} while leaving label quality uncertain~\cite{li2024abdomenatlas} due to sparse verification and the assumption that model uncertainty perfectly correlates with actual error. This raises a critical research question: How do these inherent annotation errors influence the performance of segmentation models trained on such labels? 

\noindent To address this question, we simulate the construction of modern large-scale datasets by replacing high-quality ground-truth labels with predictions from a diverse suite of models, including nnU-Net~\cite{isensee2021nnu} and several medical foundation models~\cite{ma2024segment,wasserthal2023totalsegmentator,huang2023stu}.
Comparing predictions to the actual ground truth, we generate dataset variants with varying levels of label quality. We then train segmentation models on these ``noisy'' versions, thereby mimicking a scenario where researchers unknowingly train on imperfect, automatically generated labels. These models are evaluated across two common scenarios: In-dataset performance against the dataset's clean base version and pretraining on the noisy data followed by fine-tuning on an independent, high-quality dataset. These tasks reflect two standard workflows: direct deployment after training on noisy data, or large-scale pre-training followed by high-quality downstream fine-tuning. We find a significant disparity between these tasks: while label quality is critical for models used directly after training, fine-tuning largely compensates for poor-quality pretraining labels. Since superior pretraining labels do not consistently yield better fine-tuned models, we hypothesize that during pretraining, general structural knowledge is obtained rather than details.

\noindent \textbf{Contributions:} (1) We conduct a detailed study of label noise in medical segmentation across four datasets and seven labeling models, using realistic pseudo-label noise mimicking modern automated curation pipelines.
(2) We perform, to the best of our knowledge, the first benchmark isolating the relationship between pre-training label quality and final downstream performance after fine-tuning.
(3) We reveal a critical dichotomy: while high label quality is essential for in-domain deployment, fine-tuning effectively compensates for noisy pre-training labels. 
(4) We derive actionable insights for dataset creators and model developers. 

\noindent \textbf{Related Work:}
Our study contributes to the field of label-noise~\cite{shi2024survey}, where most of the related works come from the task of building robust models via the estimation of noise-confusion-matrices~\cite{schmidt2023probabilistic,sudre2019let}, regularization~\cite {islam2021spatially}, or robust losses~\cite{gonzalez2025robust}. While valuable, these studies typically aim to adapt models to noise and compare results against non-adapted baselines. In contrast, we observe that researchers often train standard models on newly released datasets without explicit noise correction. We therefore fix the model architecture and systematically vary the data to emulate this realistic setting of unknown label quality. 
Furthermore, existing studies on label-noise impact often focus on RGB modalities~\cite{Wesemeyer2021AnnotationQV}, microscopy~\cite{jimenez2024impact}, or utilize 2D models for volumetric data~\cite{heller2018imperfect,marcinkiewicz2019quantitative,vorontsov2021label}, thereby ignoring critical inter-slice relationships. The works which natively work with 3D data often focus on single, very specific structures, such as the mandibles~\cite{yu2020robustness} or the parotid glands~\cite{strijbis2024impact}, and work on small application-specific datasets. 

\noindent Existing studies share a significant limitation: noise is typically introduced synthetically via morphological operations~\cite{bruckner2024influence,strijbis2024impact} or perturbations~\cite{heller2018imperfect}, which bake prior assumptions into the error model. Others focus on inter-annotator disagreement~\cite{Wesemeyer2021AnnotationQV,yu2020robustness}, a setting ideal for manually segmented datasets, but falling short of capturing the construction pipelines of modern datasets~\cite{jaus2024towards,qu2024abdomenatlas,wasserthal2023totalsegmentator} which heavily rely on pseudolabels. 
We address these gaps by modeling noise as realistic pseudo-label errors across datasets of varying scales. Furthermore, we utilize native 3D models to preserve volumetric context and provide the first analysis of how label quality specifically impacts the pre-training phase.

\section{Methodology}

\subsection{From Labels to Pseudo-Labels}

Pseudo-labeled datasets are defined as derivatives of a base dataset $D$ as
\begin{equation}
    D_{g}=\{(X^i,g(X^i))\}_{i=1}^N
    \label{Eq: Pseudolabel Generation}
\end{equation}
where $g$ is a neural network, capable of generating predictions $g(X^i)=\hat{Y}^i_g$ based on the input $X^i$. Within this work, we consider a set of pseudo-label generators $\mathcal{G} = $\{nnU-Net~\cite{isensee2021nnu}, MedSAM~\cite{ma2024segment}, 
STU-Net\textsubscript{\{small, base, large, huge\}}~\cite{huang2023stu}, TotalSegmentator~\cite{wasserthal2023totalsegmentator}\}, 
allowing us to generate a total of $7$ different pseudo-label version (one for each $g \in \mathcal{G}$) per base dataset. We define $\mathcal{G}^+ = \mathcal{G} \cup \{\text{base}\}$ to include the original dataset labels.

\noindent\textbf{Pseudo-Label Generators:} We employ three categories of generators, each representing a distinct family of medical segmentation models. The first category comprises standard trainable networks, represented by nnU-Net~\cite{isensee2021nnu}, being one of the most popular and successful segmentation frameworks. We use five-fold cross-validation, training separate models for each fold of the base datasets $D_{\text{base}}$. Each model is trained on four folds and predicts the respective fifth fold, which we keep as pseudolabels. Aggregating these out-of-fold predictions yields $D_{\text{nnU-Net}}$, now composed entirely of predicted labels. 

\begin{figure}[t]
  \centering
  \begin{subfigure}{0.49\columnwidth}
    \centering
    \includegraphics[width=\linewidth,trim=0 0 0 22,clip]{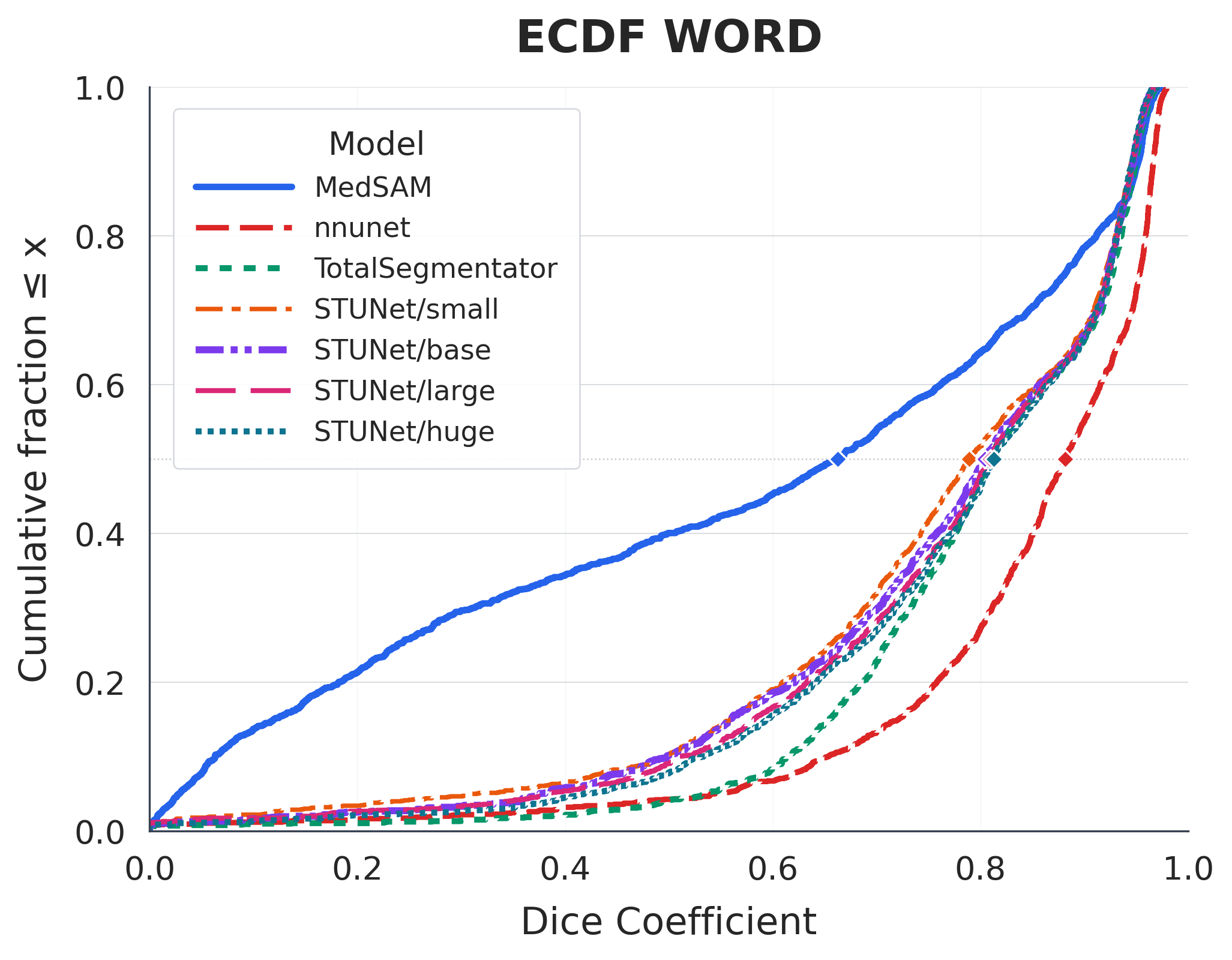}
  \end{subfigure}\hfill
  \begin{subfigure}{0.49\columnwidth}
    \centering
    \includegraphics[width=\linewidth,trim=0 0 0 22,clip]{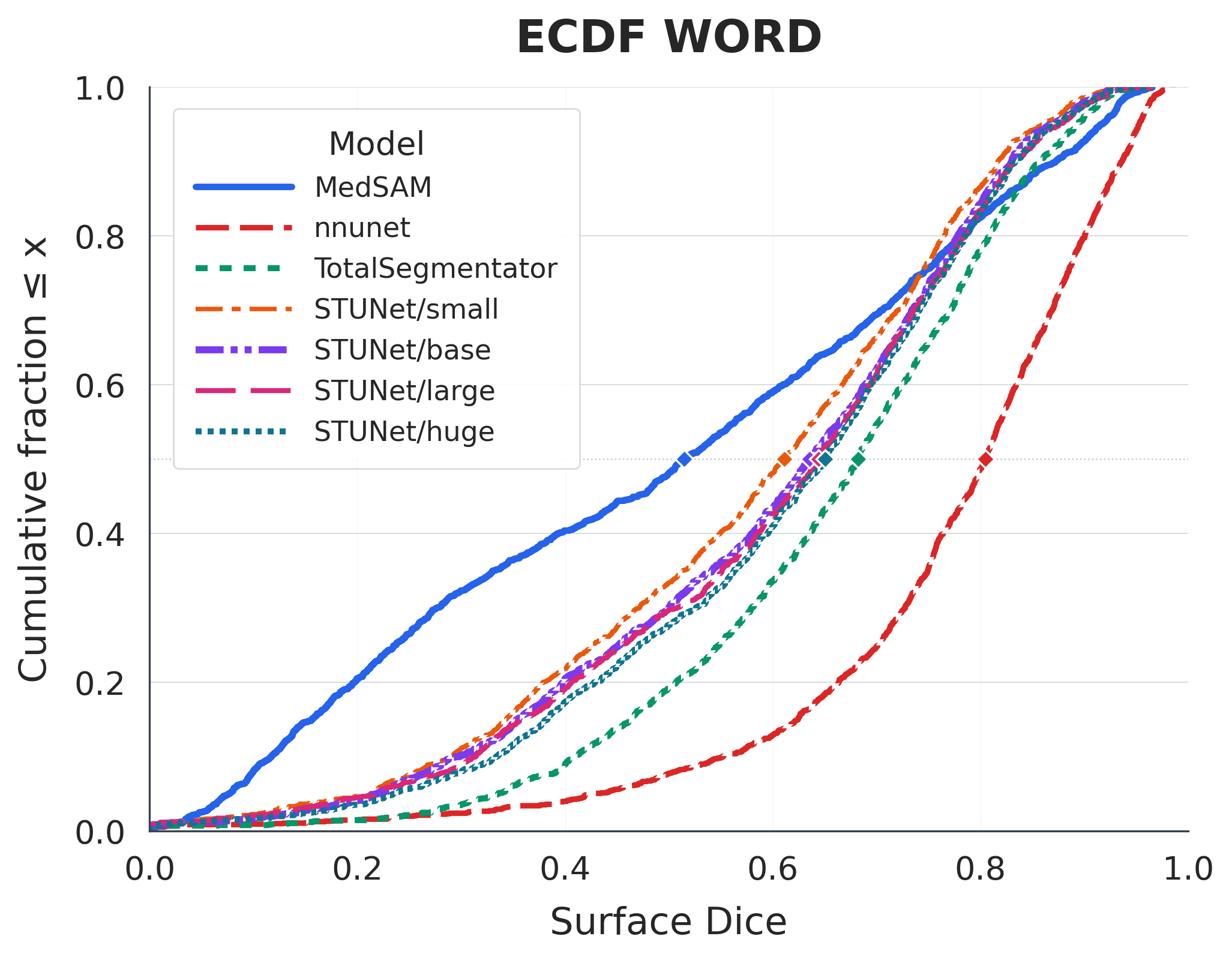}
  \end{subfigure}
  \caption{Empirical cumulative distribution functions (ECDFs) of pseudo-label quality for base dataset Word. Each curve shows the proportion of predictions whose dice/surface-dice value is at most x, averaged over all anatomical classes. A curve shifted further to the right indicates higher overall segmentation quality. nnU-Net pseudo-labels consistently dominate, while MedSAM exhibits the lowest quality. Intermediate positions are rather close but are typically ordered from STUNet, small to huge, with TotalSegmentator in between.}
  \label{fig:ecdf_word}
\end{figure}

The second category comprises ``out-of-the-box'' foundation models: TotalSegmentator~\cite{wasserthal2023totalsegmentator} and the STU-Net family~\cite{huang2023stu}. TotalSegmentator is a widely used segmentation model that is capable of robustly segmenting large parts of the human anatomy. STU-Net is a family of U-Net~\cite{ronneberger2015u} based networks which differ in parameter sizes. The authors find that scaling model parameters tends to slightly but consistently improve segmentation results. For our study, this poses an opportunity, as the small differences in performance that the different STU-Net variants yield are an ideal simulation of small quality fluctuations of annotations; this behavior is explored in~\Cref{fig:ecdf_word} for the Word~\cite{luo2022word} dataset, with other datasets behaving similarly. 


Finally, we use the interactive foundation model MedSAM~\cite{ma2024segment} serving as a bridge between the in-domain dataset generators being perfectly able to capture the annotation style of the dataset, and non-interactive models whose predictions cannot be altered. For MedSAM, we slice volumetric images and generate bounding boxes around ground-truth segmentation masks to prompt the model, allowing it to adapt to the original labels without requiring training. 

\noindent\textbf{Selection of base Datasets:} As base datasets $D$, we select 
WORD~\cite{luo2022word} (100 cases, 16 organs), AMOS~\cite{ji2022amos} (200 cases, 15 organs), CT-1K~\cite{ma2021abdomenct} (1,000 cases, 4 organs), and AbdomenAtlas~\cite{qu2024abdomenatlas,li2024how} (5,200 cases, 9 organs). These abdominal datasets were selected to represent a range of popular benchmarks and scales, facilitating a systematic study of label quality across varying organs and data magnitudes.

\setlength{\tabcolsep}{5pt}
\begin{table}
\centering \footnotesize
\caption{Overview of the quality of different pseudo-label dataset variants}
\begin{tabular}{l|l|c|c|c|c}
\toprule
 &  & Word & Amos & CT1K & Abd. Atlas \\
\midrule
\multirow{2}{*}{nnUnet}  
 & Dice       & 83.7$\pm$12.8 & 89.2$\pm$12.6 & 95.2$\pm$4.3 & 90.6$\pm$13.7 \\
 & Surf. Dice & 76.5$\pm$8.7  & 85.2$\pm$9.2  & 88.9$\pm$9.4 & 84.4$\pm$9.2 \\
\midrule
\multirow{2}{*}{MedSAM}  
 & Dice       & 56.8$\pm$18.6 & 59.9$\pm$23.7 & 72.9$\pm$13.6 & 39.4$\pm$20.5 \\
 & Surf. Dice & 49.6$\pm$7.0  & 51.9$\pm$7.2  & 57.8$\pm$7.6  & 24.8$\pm$5.0 \\
\midrule
\multirow{2}{*}{TotalS}  
 & Dice       & 79.3$\pm$10.3 & 82.9$\pm$13.7 & 91.9$\pm$5.5 & 86.6$\pm$16.4 \\
 & Surf. Dice & 65.4$\pm$8.1  & 71.6$\pm$8.5  & 80.3$\pm$9.0 & 76.1$\pm$8.7 \\
\midrule
\multirow{2}{*}{STU S}   
 & Dice       & 74.8$\pm$13.6 & 80.9$\pm$15.5 & 91.5$\pm$6.5 & 83.2$\pm$19.9 \\
 & Surf. Dice & 57.6$\pm$7.6  & 66.8$\pm$8.2  & 78.6$\pm$8.9 & 70.0$\pm$8.4 \\
\midrule
\multirow{2}{*}{STU B}   
 & Dice       & 75.9$\pm$12.5 & 82.1$\pm$15.5 & 92.0$\pm$6.4 & 84.7$\pm$18.3 \\
 & Surf. Dice & 59.5$\pm$7.7  & 66.7$\pm$8.3  & 80.1$\pm$8.9 & 72.7$\pm$8.5 \\
\midrule
\multirow{2}{*}{STU L}   
 & Dice       & 76.5$\pm$12.9 & 82.7$\pm$15.1 & 92.2$\pm$6.0 & 85.2$\pm$18.2 \\
 & Surf. Dice & 60.0$\pm$7.7  & 69.6$\pm$8.3  & 80.6$\pm$9.0 & 73.8$\pm$8.6 \\
\midrule
\multirow{2}{*}{STU H}   
 & Dice       & 77.2$\pm$12.1 & 83.0$\pm$14.6 & 92.2$\pm$5.8 & 85.8$\pm$17.4 \\
 & Surf. Dice & 60.9$\pm$7.8  & 69.7$\pm$8.3  & 80.8$\pm$9.0 & 74.4$\pm$8.6 \\
\bottomrule
\end{tabular}
\label{Tab: pseudo-label quality}
\end{table}

\subsection{Assessing the Quality of the Pseudo-Label datasets}
\label{SS: Assessing the Quality of the Pseudo-Label datasets}
For each base dataset $D$, we compute the different pseudo-label variants as outlined in~\Cref{Eq: Pseudolabel Generation}. Following~\cite{ma2021abdomenct}, we measure label quality via Dice and Surface Dice to assess how well each pseudo-label dataset (e.g., $Word_{\text{nnUnet}}$) matches its original labels ($Word_{\text{base}}$). We report the results in~\Cref{Tab: pseudo-label quality} by averaging first across the classes, followed by averaging across samples for both Dice and Surface Dice. To aggregate the organ-level standard deviations, we compute the root-mean-square across the images in the dataset.  
Ensuring a fair comparison, we only consider anatomical structures that all generators $g\in \mathcal{G}$ are capable of predicting and ignore the others in all averages. This is necessary since the non-interactive pseudo-label predictors only offer a fixed set of labels that cannot be altered. These are, however, only minimal adjustments affecting the class \textit{Prostate} in Amos and \textit{Rectum} in Word, which are excluded from the averages alongside the \textit{Background} class. 

Upon inspection of the results in~\Cref{Tab: pseudo-label quality} and \Cref{fig:ecdf_word}, we report the following findings: (1) There is a clear improvement pattern from STU S → B → L → H, though the gains between models become smaller with increasing model size, confirming the findings of~\cite{huang2023stu}. We show this behavior in more detail in~\Cref{fig:ecdf_word} for the Word dataset, with the other base datasets behaving similarly. 
(2) nnUNet outperforms on most datasets, achieving Dice scores from 83.7\% to 95.2\%, benefiting from direct training on each dataset’s annotation style. 
(3) MedSAM, as the candidate for an interactive foundation model, generally has a lower quality of predictions, which can be attributed to the model natively operating in 2D, making it unable to capture the volumetric nature of the data. Stacking its 2D predictions back into 3D introduces artifacts besides the in-slice errors. While there are approaches to adapt SAM-based models to 3D (e.g. via adapters~\cite{wu2025medical}), this is not the goal of this work, and we deliberately treat the reduced label quality of MedSAM's predictions as a proxy for low-quality slice-based labels, thereby spanning a realistic spectrum of pseudo-label quality from lower to higher regimes.

\section{Experiments and Results}
We evaluate the impact of dataset quality across two scenarios: In-domain evaluation and pre-training suitability.

\noindent\textbf{Training Setup:} We chose DynUnet~\cite{isensee2019dynunet} as our model of choice, as it offers a flexible and well-proven architecture serving as the workhorse of the nnU-Net~\cite{isensee2021nnu} framework. 
To mitigate any influence on model training except for the data, we use a completely deterministic training for 1000 epochs, where each epoch consists of exactly 250 mini-batches, thereby oversampling the dataset if needed. We generally follow the nnUnet training setup by resampling images to their respective datasets' median resolution before extracting patches. For the in-domain evaluation setup, we choose patch sizes to be specified by each dataset $D$ and remain constant across the different versions $D_{g}\forall g\in\mathcal{G}^+$. The network weights are updated using the AdamW~\cite{loshchilov2018decoupled} optimizer over a cosine-annealing learning rate scheduler starting from a base learning rate of $1e-3$. We leverage random 80/20 splits, setting aside 80\% of the data for training and 20\% for evaluation. The splits are consistent across all versions of $D$. 

\begin{figure}
    \centering    \includegraphics[width=\linewidth]{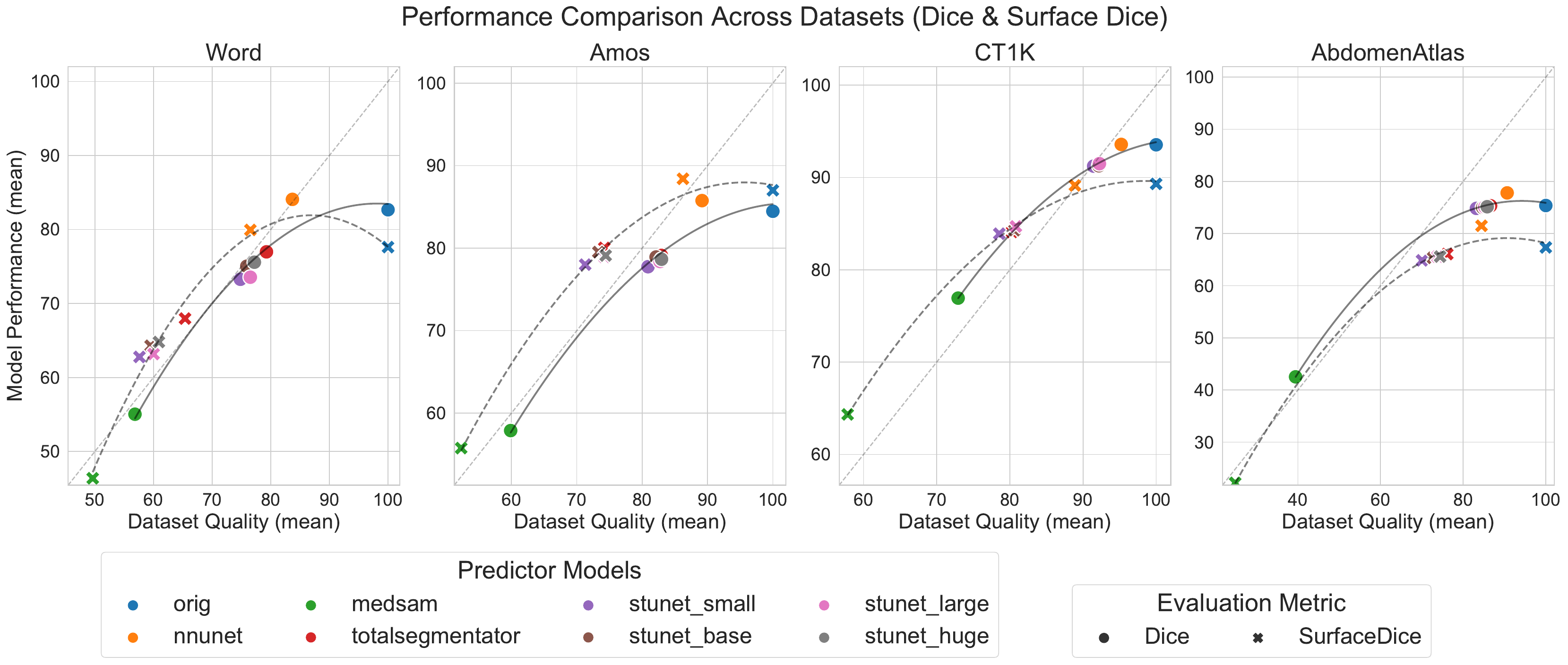}
    \caption{In-Domain evaluation results for Dice and Surface Dice. We include dashed $y=x$ for reference and perform a second degree polynomial regression on the Dice data (solid line) and Surface Dice (dashed line).}
    \label{fig:in-domain-summary}
\end{figure}

\subsection{In-Domain Evaluation}
For the task of in-domain evaluation, we consider the setup, where a model $f_{D_{g}},g\in \mathcal{G}^+$, that was trained on $D_g$ is evaluated against the testset of $D_{base}$. This scenario reflects the setting in which the target structures are predefined, yet no manual ground-truth annotations are accessible for training, such that models may be trained solely on pseudo-labels and are directly evaluated on the original reference distribution. 
We report the results in ~\Cref{fig:in-domain-summary} and summarize our findings as follows: (1) Overall, we see a strong positive correlation between dataset quality and model performance. As the quality of the dataset increases, there is a distinct rise in the resulting model performance. (2) The relationship between label quality and performance is non-linear, showing diminishing returns at the higher end of the spectrum. As the dataset quality approaches 100, the performance curves begin to plateau, falling below the diagonal reference line. This indicates that minor improvements in already high-quality labels do not necessarily translate to proportional performance gains.
(3) We observe a scaling effect where dataset volume appears to compensate for label quality. In smaller datasets such as Word or Amos, improving labels of already high quality is still reflected in considerable gains, whereas in the massive AbdomenAtlas dataset (5k cases), performance plateaus early, suggesting that sheer data quantity compensates for imperfect labels. Consequently, the "quality-performance" correlation is most critical for smaller cohorts or low-quality datasets.
(4) Among the intermediate pseudo-label generators (STU-Net variants and TotalSegmentator), incremental improvements in label quality do not guarantee monotonic performance gains, as minor enhancements may not provide enough signal to make a clear difference.
(5) Boundary accuracy (Surface Dice) consistently lags behind volumetric overlap (Dice) across all quality levels, indicating that precise surface segmentation remains more challenging than general overlap.


\begin{figure}
    \centering
    
    \begin{subfigure}{\linewidth}
    \begin{subfigure}{0.49\linewidth}
        \centering
        \includegraphics[width=\linewidth,trim=0 60 0 90,clip]{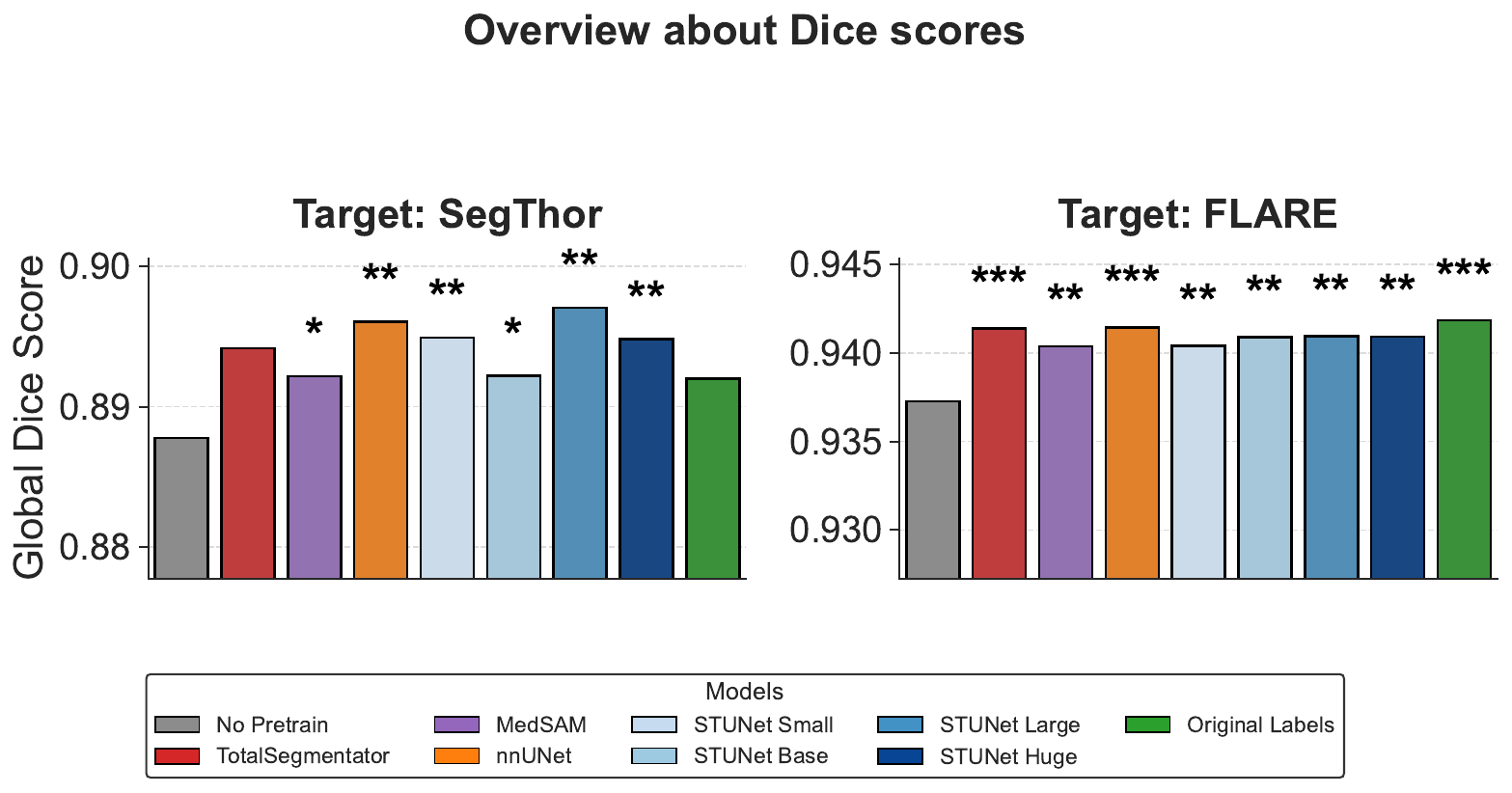}
    \end{subfigure}
    \hfill
    \begin{subfigure}{0.49\linewidth}
        \centering
        \includegraphics[width=\linewidth,trim=0 60 0 90,clip]{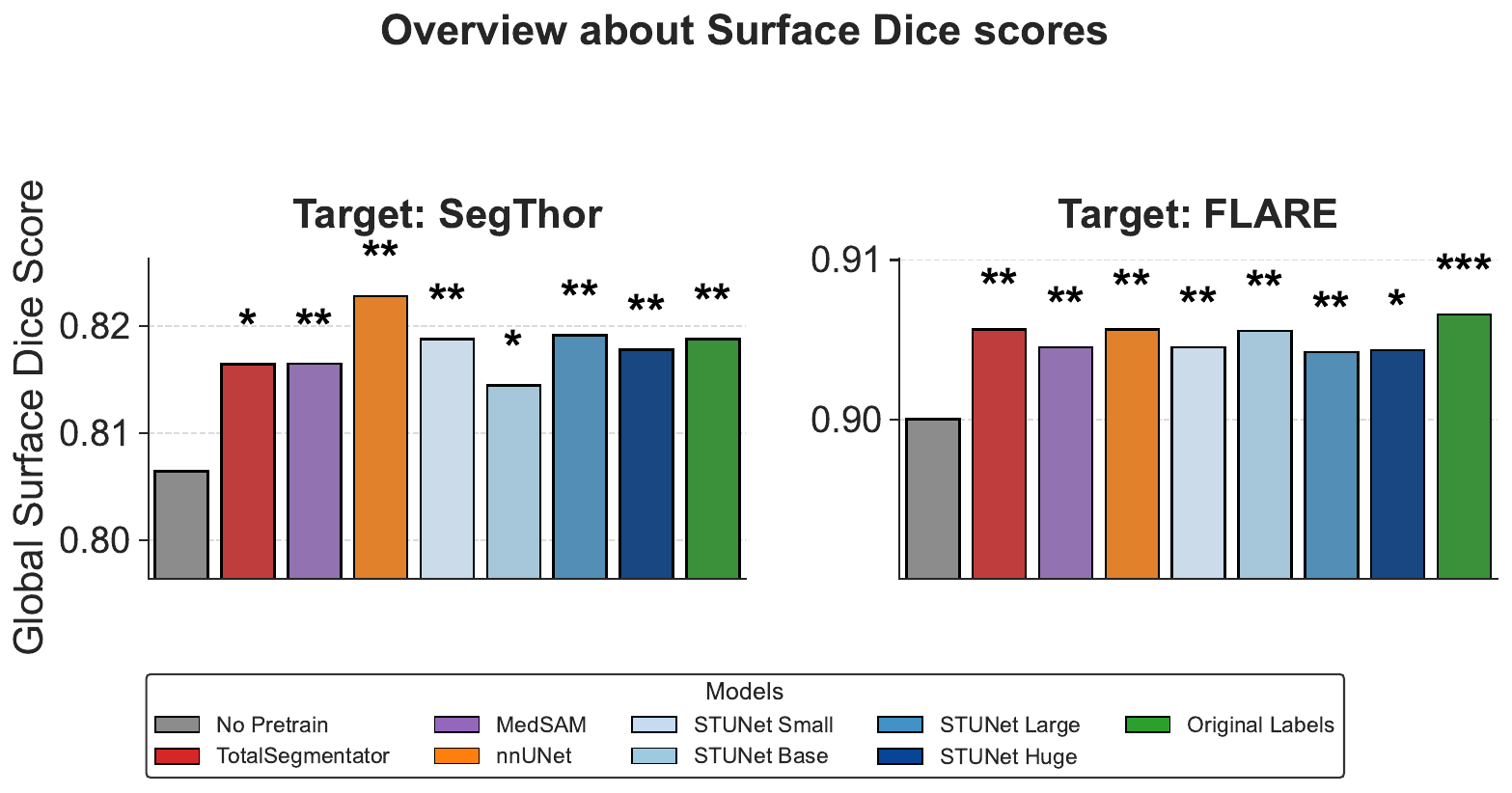}
    \end{subfigure}
    \includegraphics[width=\linewidth,trim=0 0 0 330,clip]{Latex/Images/miccai_global_sdice_valid_stats.pdf}
    \caption{Dice and Surface Dice scores on the SegThor and FLARE benchmarks. Each bar represents the average scores across different base datasets and individual organs. Stars above a bar indicate that the model significantly outperforms the \textit{No Pretrain} setting according to a one-sided Wilcoxon signed-rank test (*: p < 0.1, **: p < 0.05, ***: p < 0.01)}
    \end{subfigure}
    \vspace{0.5em}
    
    \begin{subfigure}{\linewidth}
        \centering
        \includegraphics[width=\linewidth,trim=0 60 0 0,clip]{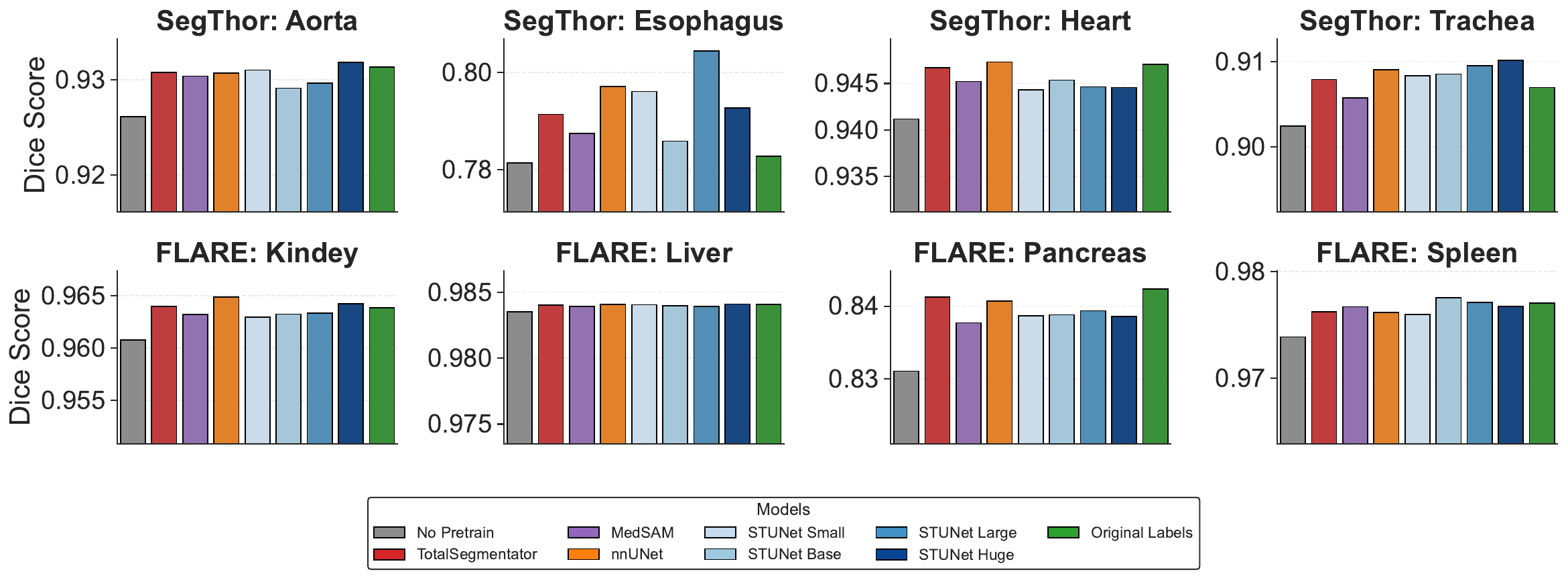}
    \end{subfigure}
    
    \vspace{0.5em}
    
    \begin{subfigure}{\linewidth}
        \centering
        \includegraphics[width=\linewidth,trim=0 0 0 330,clip]{Latex/Images/miccai_global_sdice_valid_stats.pdf}
    
    \caption{Dice scores of individual organ masks on SegThor and Flare. Each bar represents the averaged organ-mask scores across the different base-datasets.}
    \end{subfigure}
    
    \vspace{0.5em}
    
    \begin{subfigure}{\linewidth}
        \centering
        \includegraphics[width=1\linewidth,trim=0 0 0 0,clip]{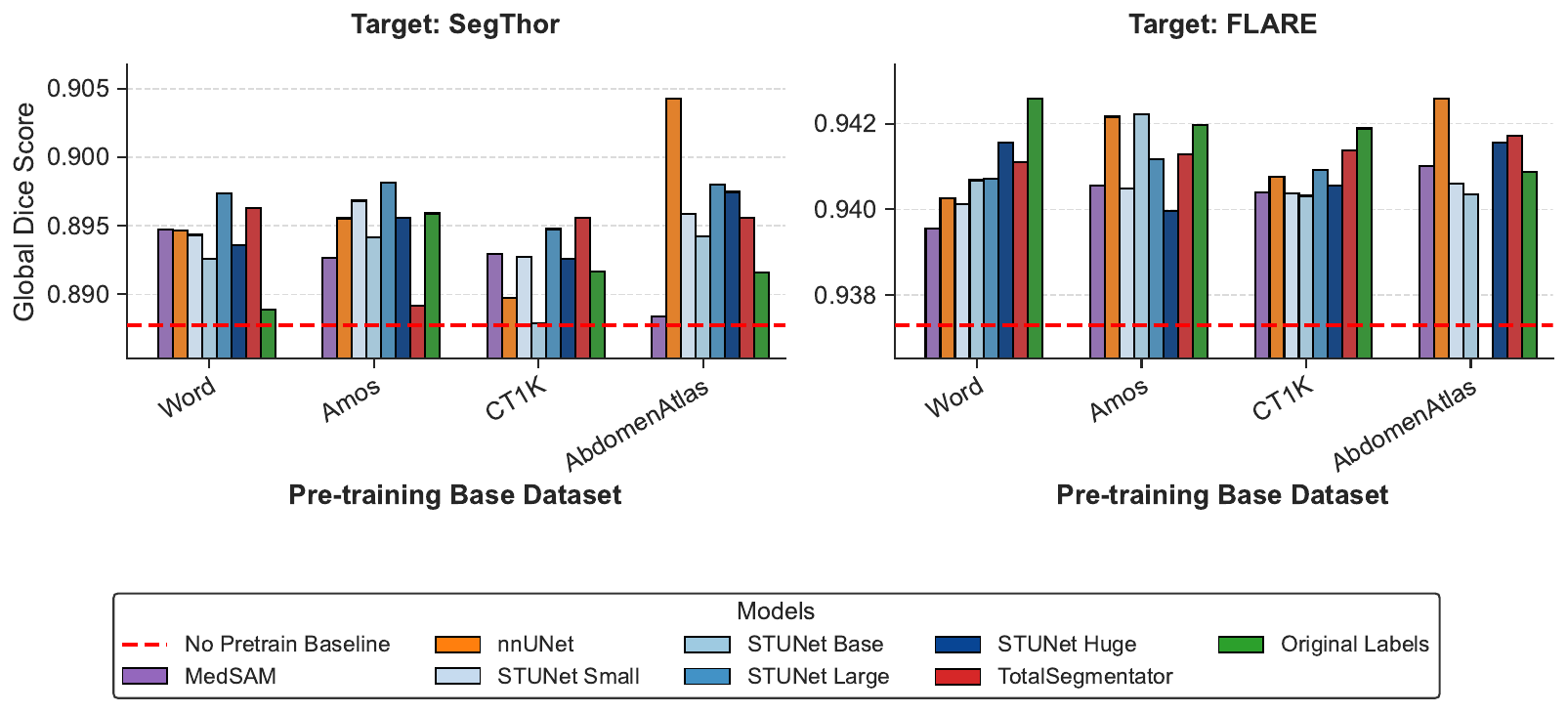}
    
    
     \caption{Dice scores of averaged organ mask on SegThor and Flare datasets, highlighting details on the effects of the pre-training base-datasets and label-generators.}
    \end{subfigure}
    
    \caption{Pre-training evaluation results: we report global and organ-specific segmentation performances on the SegThor and FLARE datasets, alongside details regarding the relationship of base datasets and label generators.}
    \label{fig:miccai_results}
\end{figure}

\subsection{Pre-training Suitability}
We now turn towards the scenario of pre-training models on the generated noisy dataset before fine-tuning them on one of two different clean target datasets: SegThor~\cite{lambert2020segthor} and Flare~\cite{MedIA-FLARE21}. While the core training procedure remains consistent, we adapt the patch size and normalization schemes to match the target datasets. For each pre-trained model $f_{D_g},g \in \mathcal{G}^+$, we initialize the fine-tuning stage with the pre-trained weights and repeat the training process on the target data. The resulting models are evaluated on the respective test sets of the target datasets, with findings detailed across three subplots in \Cref{fig:miccai_results}. In the first subplot, we present global averaged scores aggregated across all organs and base datasets (e.g., Word, Amos,...). We find that for the vast majority of pretraining settings, we outperform a non-pretrained baseline significantly, as indicated by a one-sided Wilcoxon signed-rank test. We verified these gains are data-driven, as extending the baseline training from 1k to 2k steps (matching the total pre-training and fine-tuning duration) yielded no performance improvement. More importantly, we find that the label quality of the pre-training datasets no longer seems to be of concern. This is most evident in the MedSAM results: despite the poor label quality that hindered the in-domain performance of $f_{D_{MedSAM}}$ (\Cref{fig:in-domain-summary}), MedSAM remains highly effective as a pre-training source. Models trained on MedSAM labels significantly outperform the non-pretrained baseline and achieve scores comparable to models pretrained on datasets with substantially higher label quality. This observation is further supported by the organ-level scores in subplot (b). While certain structures, such as the pancreas and trachea, exhibit greater gains from pre-training compared to other labels, such as the liver, these improvements are not primarily driven by the label quality of the pre-training dataset. Across diverse anatomical structures, the performance boosts remains consistent regardless of whether the pre-training labels were high-quality or noisy. Subfigure (c) details the effect of pre-training dataset selection on downstream performance. Interestingly, under a fixed compute budget (the default for nnU-Net style training), dataset size does not appear to have a major impact. Models pre-trained on Amos or Word, which are significantly smaller than AbdomenAtlas, do not seem to learn less meaningful representations. 

\section{Conclusion}
We present the first large-scale study isolating the impact of label quality and dataset size across medical image segmentation paradigms. Our results show that while high-quality annotations are essential for models deployed directly after training, these requirements largely disappear in pretraining settings. Fine-tuning compensates for noisy pretraining labels, suggesting that pretraining primarily transfers general structural knowledge rather than precise details. We derive two actionable insights: (i) For dataset creators targeting massive pretraining datasets, extensive expert refinement may not be cost-effective. (ii) For model developers, simple pseudo-labeling of unlabeled image collections can suffice to learn transferable representations.

\begin{credits}
\subsubsection{\ackname} The present contribution is supported by the Helmholtz Association under the joint research school “HIDSS4Health - Helmholtz Information and Data Science School for Health. 
This work was performed on the HoreKa supercomputer funded by the Ministry of Science, Research and the Arts Baden-Württemberg and by the Federal Ministry of Education and Research.
This work was supported by funding from the pilot program Core-Informatics of the Helmholtz Association (HGF).

\subsubsection{\discintname}
The authors have no competing interests to declare that are relevant to the content of this article. 
\end{credits}


\bibliographystyle{splncs04}
\bibliography{bibliography}
\end{document}